%% file: naacl2021.tex
\pdfoutput=1

\documentclass[11pt]{article}

\usepackage{naacl2021}

\usepackage{times}
\usepackage{latexsym}

\usepackage[T1]{fontenc}

\usepackage[utf8]{inputenc}

\usepackage{microtype}

\input{packages}

\input{macros}

\title{Reading and Acting while Blindfolded: \\ The Need for Semantics in Text Game Agents}

\author{{Shunyu Yao$^\dagger$\thanks{~Work partly done during internship at Microsoft Research. Project page: \url{https://blindfolded.cs.princeton.edu}} \quad Karthik Narasimhan$^\dagger$ \quad Matthew Hausknecht$^\ddagger$} \\
$^\dagger$Princeton University \quad\quad  $^\ddagger$Microsoft Research \\
\texttt{\{shunyuy, karthikn\}@princeton.edu} \\
\texttt{matthew.hausknecht@microsoft.com} \\
}

\begin{document}
\maketitle
\input{text/abstract}
\input{text/intro}

\input{text/models}

\input{text/results}

\input{text/discussion}

\section*{Ethical Considerations}
Autonomous decision-making agents are potentially impactful in our society, and it is of great ethical consideration to make sure their understanding of the world and their objectives align with humans. Humans use natural language to convey and understand concepts as well as inform decisions, and in this work we investigate whether autonomous agents leverage language semantics similarly to humans in the environment of text-based games. Our findings suggest that the current generation of agents optimized for reinforcement learning objectives might not exhibit human-like language understanding, a phenomenon we should pay attention to and further study.

\section*{Acknowledgements}
We appreciate helpful suggestions from anonymous reviewers as well as members of the Princeton NLP Group and MSR RL Group.
\bibliography{main}
\bibliographystyle{acl_natbib}

\end{document}

%% file: packages.tex
\usepackage{color}
\usepackage{algorithm}
\usepackage[noend]{algpseudocode}
\usepackage[utf8]{inputenc} %
\usepackage[T1]{fontenc}    %
\usepackage{hyperref}       %
\usepackage{url}            %
\usepackage{booktabs}       %
\usepackage{amsfonts}       %
\usepackage{nicefrac}       %
\usepackage{microtype}      %
\usepackage[linewidth=1pt]{mdframed}
\usepackage{tikz-dependency}
\usepackage{amsmath,amsfonts,bm}
\usepackage{graphicx}
\usepackage{algorithmicx}
\usepackage[noend]{algpseudocode}
\usepackage{booktabs}
\usepackage[percent]{overpic}
\usepackage{subcaption}
\usepackage{siunitx}
\usepackage{wrapfig}
\usepackage{listings}
\usepackage{multirow}
\usepackage[page]{appendix}

\usepackage{hyperref}
\usepackage{url}
\NewEnviron{elaboration}{
\par
\begin{tikzpicture}
\node[rectangle,minimum width=\textwidth] (m) {\begin{minipage}{.98\textwidth}\BODY\end{minipage}};
\draw[dashed] (m.south west) rectangle (m.north east);
\end{tikzpicture}
}
\usepackage{enumitem}

\usepackage[export]{adjustbox}

%% file: macros.tex
\newcolumntype{L}[1]{>{\raggedright\let\newline\\\arraybackslash\hspace{0pt}}m{#1}}
\newcolumntype{C}[1]{>{\centering\let\newline\\\arraybackslash\hspace{0pt}}m{#1}}
\newcolumntype{R}[1]{>{\raggedleft\let\newline\\\arraybackslash\hspace{0pt}}m{#1}}

\newcommand{\fig}[1]{Figure~\ref{#1}}
\newcommand{\tbl}[1]{Table~\ref{#1}}

\newcommand{\ignorethis}[1]{}

\newcommand{\ignore}[1]{}

\makeatletter
\DeclareRobustCommand\onedot{\futurelet\@let@token\@onedot}
\def\@onedot{\ifx\@let@token.\else.\null\fi\xspace}

\makeatother

%% file: text/abstract.tex
\begin{abstract}
  Text-based games simulate worlds and interact with players using natural language. Recent work has used them as a testbed for autonomous language-understanding agents, with the motivation being that understanding the meanings of words or \textit{semantics} is a key component of how humans understand, reason, and act in these worlds. 
  However, it remains unclear to what extent artificial agents utilize semantic understanding of the text. To this end, we perform experiments to systematically reduce the amount of semantic information available to a learning agent. 
  Surprisingly, we find that an agent is capable of achieving high scores even in the complete absence of language semantics, indicating that the currently popular experimental setup and models may be poorly designed to understand and leverage game texts. To remedy this deficiency, we propose an inverse dynamics decoder to regularize the representation space and encourage exploration, which shows improved performance on several games including \textsc{Zork I}. We discuss the implications of our findings for designing future agents with stronger semantic understanding.
  
\end{abstract}

%% file: text/intro.tex
\section{Introduction}

\input{figText/teaser}

Text adventure games such as \textsc{Zork I} (\fig{fig:teaser} (a)) have been a testbed for developing autonomous agents that operate using natural language. Since interactions in these games (input observations, action commands) are through text, the ability to understand and use language is deemed necessary and critical to progress through such games. Previous work has deployed a spectrum of methods for language processing in this domain, including word vectors~\cite{fulda17}, recurrent neural networks~\cite{narasimhan15,hausknecht19colossal},  pre-trained language models~\cite{yao2020calm}, open-domain question answering systems~\cite{ammanabrolu2020avoid}, knowledge graphs~\cite{ammanabrolu2020graph,ammanabrolu2020avoid,adhikari2020learning}, and reading comprehension systems~\cite{guo2020interactive}.

Meanwhile, most of these models operate under the reinforcement learning (RL) framework, where the agent explores the same environment in repeated episodes, learning a value function or policy to maximize game score. From this perspective, text games are just special instances of a partially observable Markov decision process (POMDP) $(S, T, A, O, R, \gamma)$, where players issue text actions $a \in A$, receive text observations $o \in O$ and scalar rewards $r = R(s, a)$, and the underlying game state $s \in S$ is updated by transition $s' = T(s, a)$. 

However, what distinguishes these games from other POMDPs is the fact that the actions and observations are in language space $L$. Therefore, a certain level of decipherable \emph{semantics} is attached to text observations $o \in O \subset L$ and actions $a \in A \subset L$. Ideally, these texts not only serve as observation or action \emph{identifiers}, but also provide clues about the latent transition function $T$ and reward function $R$. For example, issuing an action ``jump'' based on an observation ``on the cliff'' would likely yield a subsequent observation such as ``you are killed'' along with a negative reward. Human players often rely on their understanding of language semantics to inform their choices, even on games they have never played before, while replacing texts with non-semantic identifiers such as their corresponding hashes (\fig{fig:teaser} (c)) would likely render games unplayable for people. 
However, would this type of transformation affect current RL agents for such games? In this paper, we ask the following question: 
\emph{To what extent do current reinforcement learning agents leverage semantics in text-based games?}

To shed light on this question, we investigate the Deep Reinforcement Relevance Network (DRRN)~\cite{he2015deep}, a top-performing RL model that uses gated recurrent units (GRU)~\cite{cho-etal-2014-properties} to encode texts.  We conduct three experiments on a set of interactive fiction games from the Jericho benchmark~\cite{hausknecht19colossal} to probe the effect of different semantic representations on the functioning of DRRN. These include (1) using just a location phrase as the input observation (\fig{fig:teaser} (b)), (2) hashing text observations and actions (\fig{fig:teaser} (c)), and (3) regularizing vector representations using an auxiliary inverse dynamics loss. While reducing observations to location phrases leads to decreased scores and enforcing inverse dynamics decoding leads to increased scores on some games, hashing texts to break semantics surprisingly matches or even outperforms the baseline DRRN on almost all games considered. This implies current RL agents for text-based games might not be sufficiently leveraging the semantic structure of game texts to learn good policies, and points to the need for developing better experiment setups and agents that have a finer grasp of natural language.

%% file: figText/teaser.tex
\begin{figure*}[t]
\small
\begin{subfigure}{.6\textwidth}
\begin{mdframed}
\begin{center}
(a) \textsc{Zork I}
\end{center}

\emph{Observation 21:} 
You are in the living room. There is a doorway to the east, a wooden door with strange gothic lettering to the west, which appears to be nailed shut, a trophy case, and a large oriental rug in the center of the room. You are carrying: A brass lantern \dots
\begin{flushleft}
\underline{\emph{Action 21}}: 
move rug
\end{flushleft}
\emph{Observation 22}: 
With a great effort, the rug is moved to one side of the room, revealing the dusty cover of a closed trap door... Living room... You are carrying: ...

\begin{flushleft}
\underline{\emph{Action 22}}: 
open trap
\end{flushleft}
\end{mdframed}
 \end{subfigure}
\begin{subfigure}{.38\textwidth}
 \begin{mdframed}
 \begin{center}
(b) \textbf{\textsc{min-ob}}
\end{center}

\emph{Observation 21:} Living Room

\underline{\emph{Action 21}}: move rug

\emph{Observation 22}: Living Room

\underline{\emph{Action 22}}: open trap

\end{mdframed}
 \begin{mdframed}

  \begin{center}
 (c) \textbf{\textsc{hash}}
\end{center}

\emph{Observation 21:} 0x6fc

\underline{\emph{Action 21}}: 0x3a04

\emph{Observation 22}: 0x103b

\underline{\emph{Action 22}}: 0x16bb

\end{mdframed}
\end{subfigure}
 \caption{(a): Sample original gameplay from \textsc{Zork I}. (b) (c): Our proposed semantic ablations. (b) \textbf{\textsc{min-ob}} reduces observations to only the current location name, and (c) \textbf{\textsc{hash}} replaces observation and action texts by their string hash values.}
 \label{fig:teaser}
\vspace{-4mm}

\end{figure*}

%% file: text/models.tex
\section{Models}
\paragraph{DRRN Baseline} 
Our baseline RL agent DRRN~\cite{he2015deep} learns a Q-network $Q_\phi(o, a)$ parametrized by $\phi$. The model encodes the observation $o$ and each action candidate $a$ using two separate GRU encoders $f_o$ and $f_a$, and then aggregates the representations to derive the Q-value through a MLP decoder $g$:
\begin{equation}
    Q_\phi(o, a) = g(\mathrm{concat}(f_o(o), f_a(a)))
    \label{eq:drrn}
\vspace{-1.5mm}
\end{equation}
For learning $\phi$, tuples $(o, a, r, o')$ of observation, action, reward and the next observation are sampled from an experience replay buffer and the following temporal difference (TD) loss is minimized:
\begin{equation}
   \mathcal{L}_\text{{TD}}(\phi) = (r + \gamma \max_{a' \in A} Q_\phi(o', a') - Q_\phi(o, a))^2
   \label{eq:td}
\vspace{-1.5mm}
\end{equation}
During gameplay, a softmax exploration policy is used to sample an action:
\begin{equation}
    \pi_\phi(a|o) = \frac{\exp(Q_\phi(o, a))}{\sum_{a' \in A} \exp(Q_\phi(o, a'))}
    \label{eq:softmax}
\vspace{-1.5mm}
\end{equation}
Note that when the action space $A$ is large, \eqref{eq:td} and \eqref{eq:softmax} become intractable. A valid action handicap~\cite{hausknecht19colossal} or a language model~\cite{yao2020calm} can be used to generate a reduced action space for efficient exploration. For all the modifications below, we use the DRRN with the valid action handicap as our base model.

\paragraph{Reducing Semantics via Minimizing Observation (\textsc{min-ob})} Unlike other RL domains such as video games or robotics control, at each step of text games the (valid) action space is constantly changing, and it reveals useful information about the current state. For example, knowing ``unlock box'' is valid leaks the existence of a locked box. Also, sometimes action semantics indicate its value even unconditional on the state, e.g.\,``pick gold'' usually seems good. 
Given these, we minimize the observation to only a location phrase $o \mapsto \mathrm{loc}(o)$ (\fig{fig:teaser} (b)) to isolate the action semantics:
$
    Q^{\mathrm{loc}}_\phi(o, a) = g(f_o(\mathrm{loc}(o)), f_a(a)))
$.

\paragraph{Breaking Semantics via Hashing (\textsc{hash}) } GRU encoders $f_o$ and $f_a$ in the Q-network \eqref{eq:drrn} generally ensure that similar texts (e.g.\, a single word change) are given similar representations, and therefore similar values. 
To study whether such a semantics continuity is useful, we break it by hashing observation and action texts. Specifically, given a hash function from strings to integers $h: L \to \mathbb{Z}$, and a pseudo-random generator $G: \mathbb{Z} \to \mathbb{R}^d$ that turns an integer seed to a random Gaussian vector, a hashing encoder $\hat{f} = G \circ h: L \to \mathbb{R}^d$ can be composed. While $f_o$ and $f_a$ in \eqref{eq:drrn} are trainable, $\hat{f}$ is fixed throughout RL, and ensures two texts that only differ by a word would have completely different representations. In this sense, hashing breaks semantics and only serves to identify different observations and actions in an abstract MDP problem (\fig{fig:teaser} (c)):
$
    Q^{\mathrm{hash}}_\phi(o, a) = g(\hat{f}(o), \hat{f}(a))
$.
\input{figText/result}

\paragraph{Regularizing Semantics via Inverse Dynamics Decoding (\textsc{inv-dy})} The GRU representations in DRRN $f_o(o), f_a(a)$ are only optimized for the TD loss \eqref{eq:td}. As a result, text semantics can  degenerate during encoding, and the text representation might arbitrarily overfit to the Q-values. To regularize and encourage more game-related semantics to be encoded, we take inspiration from \citet{pathakICMl17curiosity} and propose an inverse dynamics auxiliary task during RL. Given representations of current and next observations $f_o(o), f_o(o')$, we use a MLP $g_{inv}$ to predict the action representation, and a GRU decoder $d$ to decode the action back to text\footnote{Directly defining an L1/L2 loss between $f_a(a)$ and $ g_{inv}(\mathrm{concat}(f_o(o), f_o(o')))$ in the representation space will collapse text representations together.}. The inverse dynamics loss is defined as
\begin{equation*}
   \mathcal{L}_\text{{inv}}(\phi, \theta) 
   =- \log p_d(a | g_{inv}(\mathrm{concat}(f_o(o), f_o(o')))
   \label{eq:inv}
\vspace{-1.5mm}
\end{equation*}
where $\theta$ denote weights of $g_{inv}$ and $d$, and $p_d(a|x)$ is the probability of decoding token sequence $a$ using GRU decoder $d$ with initial hidden state as $x$. To also regularize the action encoding, action reconstruction from $f_a$ is also used as a loss term:
\begin{equation*}
   \mathcal{L}_\text{{dec}}(\phi, \theta) 
   =- \log p_d(a | f_a(a))
   \label{eq:dec}
\vspace{-1.5mm}
\end{equation*}
And during experience replay, these two losses are optimized along with the TD loss:
\begin{equation*}
   \mathcal{L}(\phi, \theta) 
   =  \mathcal{L}_\text{{TD}}(\phi) + \lambda_{1} \mathcal{L}_\text{{inv}}(\phi, \theta) + \lambda_{2} \mathcal{L}_\text{{dec}}(\phi, \theta) 
   \label{eq:loss}
\vspace{-1.5mm}
\end{equation*}
\input{figText/transfer}
An intrinsic reward $r^+ = \mathcal{L}_\text{{inv}}(\phi, \theta)$ is also used to explore toward where the inverse dynamics is not learned well yet. All in all, the aim of \textbf{\textsc{inv-dy}} is threefold: (1) regularize both action and observation representations to avoid degeneration by decoding back to the textual domain, (2) encourage $f_o$ to encode action-relevant parts of observations, and (3) provide intrinsic motivation for exploration.

%% file: figText/result.tex
\begin{table}[t]
    \resizebox{\columnwidth}{!}{
    \begin{tabular}{r|l|lll|l}
    \midrule
\textbf{Game} & \textbf{DRRN} & \textbf{\textsc{min-ob}} & \textbf{\textsc{hash}} & \textbf{\textsc{inv-dy}} & \textbf{Max} \\
\midrule
\midrule
balances  & 10 / 10 & 10 / 10 & 10 / 10 & 10 / 10 & 51 \\
deephome   & 57 / 66 & 8.5 / 27 & 58 / 67 & 57.6 / 67 & 300 \\
detective  & 290 / 337 & 86.3 / 350 & 290 / 317 & 290 / 323 & 360 \\
dragon   & -5.0 / 6 & -5.4 / 3 & -5.0 / 7 & \textbf{-2.7} / \textbf{8} & 25 \\
enchanter  & 20 / 20 & 20 / 40 & 20 / 30 & 20 / 30 & 400 \\
inhumane   & 21.1 / 45 & 12.4 / 40 & 21.9 / 45 & 19.6 / 45 & 90 \\
library  & 15.7 / 21 & 12.8 / 21 & \textbf{17} / 21 & 16.2 / 21 & 30 \\
ludicorp   & 12.7 / 23 & 11.6 / 21 & \textbf{14.8} / 23 & 13.5 / 23 & 150 \\
omniquest  & 4.9 / 5 & 4.9 / 5 & 4.9 / 5 & 5.3 / \textbf{10} & 50 \\
pentari   & 26.5 / 45 & 21.7 / 45 & \textbf{51.9} / \textbf{60} & 37.2 / 50 & 70 \\
zork1   & 39.4 / 53 & 29 / 46 & 35.5 / 50 & \textbf{43.1} / \textbf{87} & 350 \\
zork3   & 0.4 / 4.5 & 0.0 / 4 & 0.4 / 4 & 0.4 / 4 & 7 \\
\midrule 
Avg.\,Norm & .21 / .38 & .12 / .35 & \textbf{.25} / .39 & .23 / \textbf{.40 }&   \\
\midrule

\end{tabular}
}

\caption{Final/maximum score of different models. }
\vspace{-4mm}
\label{tab:results}
\end{table}

%% file: figText/transfer.tex
\begin{figure}[t]
    \centering
    \includegraphics[width=.5\textwidth]{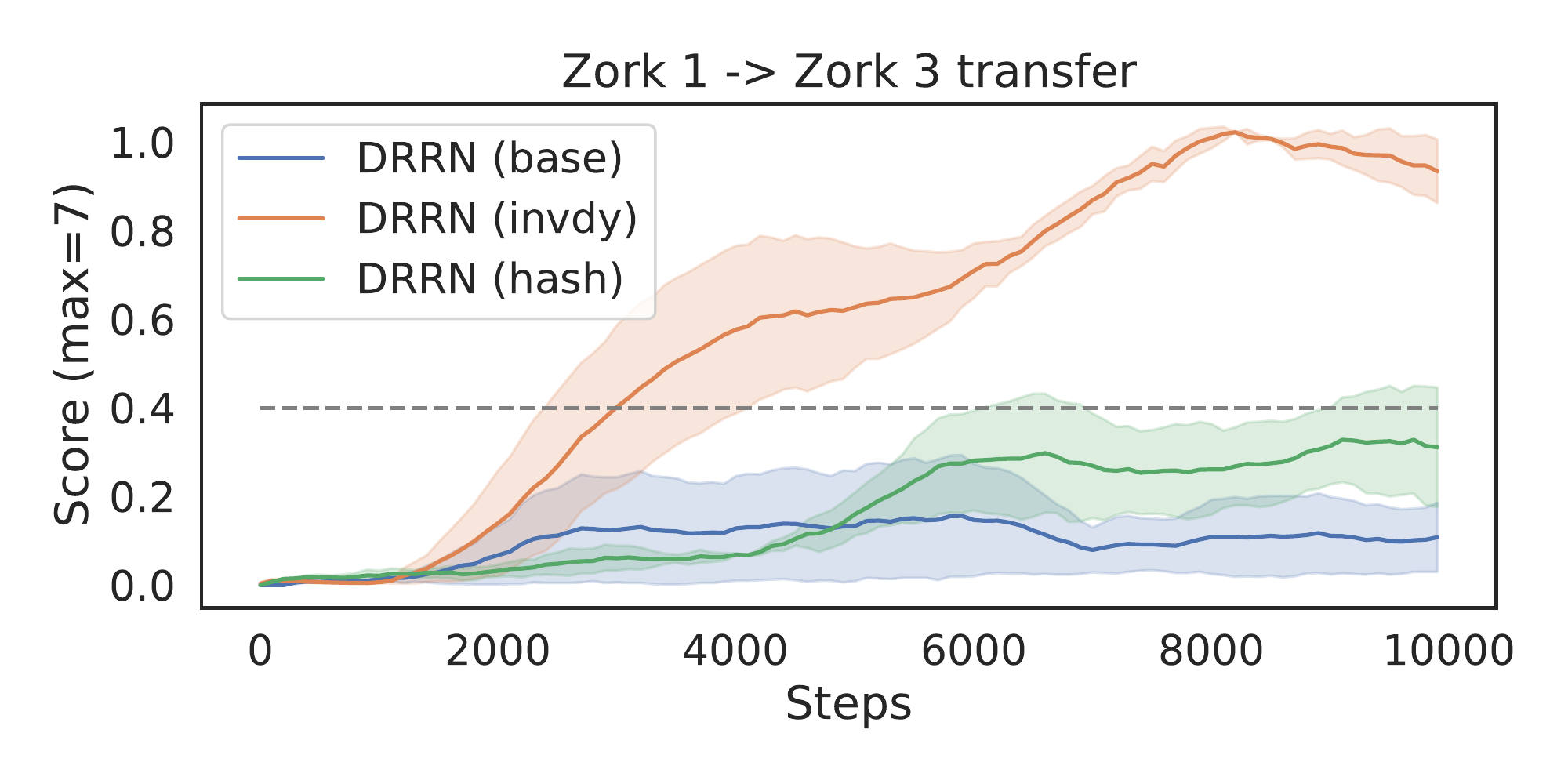}
    \caption{Transfer results from \textsc{Zork I}.}
    \label{fig:transfer}
    \vspace{-4mm}
\end{figure}

%% file: text/results.tex
\section{Results}

\paragraph{Setup} We train on 12 games\footnote{We omit games where DRRN cannot score.} from the Jericho benchmark~\cite{hausknecht19colossal}. These human-written interactive fictions are rich, complex, and diverse in semantics.\footnote{Please refer to \citet{hausknecht19colossal} for more details about these games.} For each game, we train DRRN asynchronously on 8 parallel instances of the game environment for $10^5$ steps, using a prioritized replay buffer. Following prior practice~\cite{hausknecht19colossal}, we augment observations with location and inventory descriptions by issuing the ‘look’ and ‘inventory’ commands. 
We train three independent runs for each game and report their average score. For \textbf{\textsc{hash}}, we use the Python built-in hash function to process text as a tuple of token IDs, and implement the random vector generator $G$ by seeding PyTorch with the hash value. For \textbf{\textsc{inv-dy}}, we use $\lambda_{1} = \lambda_{2} = 1$.

\input{figText/vis}

\paragraph{Scores} \tbl{tab:results} reports the final score (the average score of the final 100 episodes during training), and the maximum score seen in each game for different models. Average normalized score (raw score divided by game total score) over all games is also reported. Compared to the base DRRN, \textbf{\textsc{min-ob}} turns out to explore similar maximum scores on most games (except \textsc{Deephome} and \textsc{Dragon}), but fails to memorize the good experience and reach high episodic scores, which suggests the importance of identifying different observations using language details. Most surprisingly, \textbf{\textsc{hash}} has a lower final score than DRRN on only one game (\textsc{Zork I}), while on \textsc{Pentari} it almost doubles the DRRN final score. It is also the model with the best average normalized final score across games, which indicates that the DRRN model can perform as well without leveraging any language semantics, but instead simply by identifying different observations and actions with random vectors and memorizing the Q-values. 
Lastly, we observe on some games (\textsc{Dragon}, \textsc{Omniquest}, \textsc{Zork I}) \textbf{\textsc{inv-dy}} can explore high scores that other models cannot. Most notably, on \textsc{Zork I} the  maximum score seen is 87 (average of $54, 94, 113$), while any run of other models does not explore a score more than 55. This might indicate potential benefit of developing RL agents with more semantic representations.

\paragraph{Transfer} We also investigate if representations of different models can transfer to a new language environment, which is a potential benefit of learning natural language semantics.
So we consider the two most similar games in Jericho, \textsc{Zork I} and \textsc{Zork III}, fix the language encoders of different \textsc{Zork I} models, and re-train the Q-network on \textsc{Zork III} for 10,000 steps.
As shown in \fig{fig:transfer}, \textbf{\textsc{inv-dy}} representations can achieve a score around 1, which surpasses the best result of models trained from scratch on \textsc{Zork III} for 100,000 steps (around 0.4), showing great promise in better gameplay by leveraging language understanding from other games. \textbf{\textsc{hash}} transfer is equivalent to training from scratch as the representations are not learnt, and a score around 0.3 is achieved. Finally, DRRN representations transfer worse than \textbf{\textsc{hash}}, possibly due to overfitting to the TD loss \eqref{eq:td}.

\paragraph{{Visualizations}} Finally, we use t-SNE~\cite{maaten2008visualizing} to visualize representations of some \textsc{Zork I} walkthrough states in \fig{fig:vis}. The first 30 walkthrough states (red, score 0-45) are well experienced by the models during exploration, whereas the last 170 states (blue, score 157-350) are unseen\footnote{The rest 150 states in the middle (score 45-157) are omitted as they might be seen by some model but not others.}. We also encircle the subset of states at location `living room' for their shared semantics.

First, we note that the \textbf{\textsc{hash}} representations for living room states are scattered randomly, unlike the other two models with GRU language encoders.
Further, the base DRRN overfits to the TD loss \eqref{eq:td},  representing unseen states (blue) in a different subspace to seen states (red) without regarding their semantic similarity. 
\textbf{\textsc{ind-dy}} is able to extrapolate to unseen states and represent them similarly to seen states for their shared semantics, which may explain its better performance on this game.

\paragraph{Game stochasticity} All the above experiments were performed using a fixed game random seed for each game, following prior work~\cite{hausknecht19colossal}. To investigate if randomness in games affects our conclusions, we run one trial of each game with episode-varying random seeds\footnote{Randomness includes transition uncertainty (e.g.\,thief showing up randomly in \textsc{Zork I}) and occasional paraphrasing of text observations.}. We find the average normalized score for the base DRRN, \textbf{\textsc{hash}}, \textbf{\textsc{inv-dy}} to be all around 17\%, with performance drop mainly on three stochastic games (\textsc{Dragon}, \textsc{Zork I}, \textsc{Zork III}). Notably, the core finding that the base DRRN and \textbf{\textsc{hash}} perform similarly still holds. Intuitively, even though the Q-values would be lower overall with unexpected transitions, RL would still memorize observations and actions that lead to high Q-values.

%% file: figText/vis.tex
\begin{figure*}[ht]
    \centering
    \includegraphics[width=\textwidth]{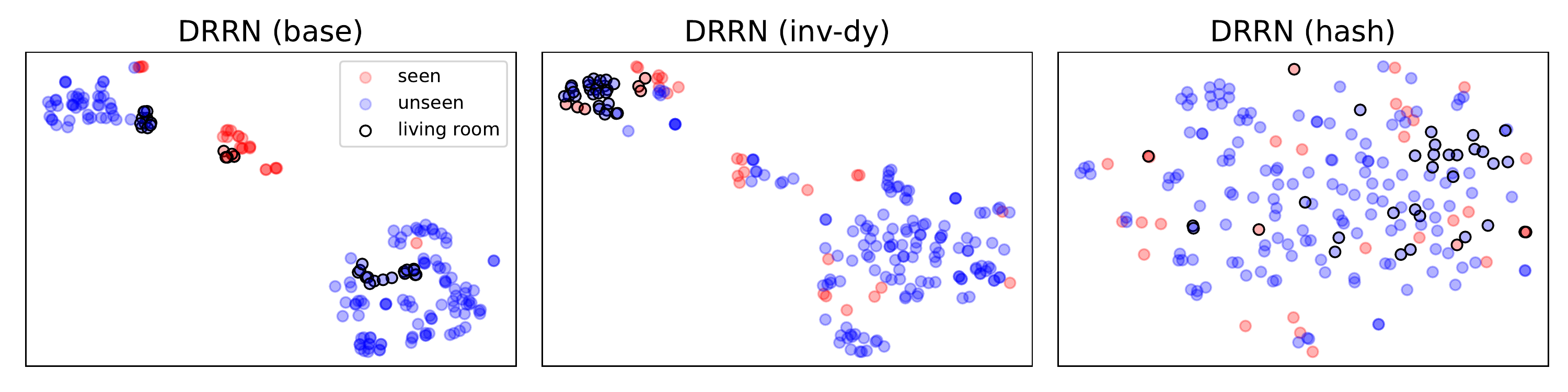}
    \caption{t-SNE visualization of seen and unseen state observations of \textsc{Zork I}. DRRN (base) represents unseen states separated from seen states while \textbf{\textsc{inv-dy}} mixes them by semantic similarity.}
    \label{fig:vis}
    \vspace{-4mm}
\end{figure*}

%% file: text/discussion.tex
\section{Discussion}

At a high level, RL agents for text-based games succeed by (1) exploring trajectories that lead to high scores, and (2) learning representations to stably reach high scores. Our experiments show that a semantics-regularized \textbf{\textsc{inv-dy}} model manages to explore higher scores on some games (\textsc{Dragon}, \textsc{Omniquest}, \textsc{Zork I}), while the \textbf{\textsc{hash}} model manages to memorize scores better on other games (\textsc{Library}, \textsc{Ludicorp}, \textsc{Pentari}) using just a fixed, random, non-semantic representation. This leads us to  hypothesize two things. First, fixed, stable representations might make  Q-learning easier. Second, it might be desirable to represent similar texts very differently for better gameplay, e.g.\,the Q-value can be much higher when a key object is mentioned, even if it only adds a few words to a long observation text. This motivates future thought into the structural vs.\,functional use of language semantics in these games.

Our findings also urge a re-thinking of the popular `RL + valid action handicap' setup for these games. On one hand, RL sets training and evaluation in the same environment, with limited text corpora, and sparse, mostly deterministic rewards as the only optimization objective. Such a combination easily results in overfitting to the reward system of a specific game (\fig{fig:transfer}), or even just a specific stage of the game (\fig{fig:vis}). On the other hand, the valid action handicap reduces the action set to a small size tractable for memorization, and reduces the language understanding challenge for the RL agent.
Thus for future research on text-based games, we advocate for more attention towards alternative setups without RL or handicaps~\cite{hausknecht2019nail,yao2020calm,yu2020deriving}. Particularly, in a `RL + no valid action handicap' setting, \emph{generating} action candidates rather than simply \emph{choosing} from a set entails more opportunities and challenges with respect to learning grounded language semantics~\cite{yao2020calm}.
Additionally, training agents on a distribution of games and evaluating them on a separate set of unseen games would require more general semantic understanding.
Semantic evaluation of these proposed paradigms is outside the scope of this paper, but we hope it will spark a productive discussion on the next steps toward building agents with stronger semantic understanding.

%% file: naacl2021.bbl
\begin{thebibliography}{14}
\expandafter\ifx\csname natexlab\endcsname\relax\def\natexlab#1{#1}\fi

\bibitem[{Adhikari et~al.(2020)Adhikari, Yuan, Côté, Zelinka, Rondeau,
  Laroche, Poupart, Tang, Trischler, and ~}]{adhikari2020learning}
Ashutosh Adhikari, Xingdi~(Eric) Yuan, Marc-Alexandre Côté, Mikulas Zelinka,
  Marc-Antoine Rondeau, Romain Laroche, Pascal Poupart, Jian Tang, Adam
  Trischler, and William L.~Hamilton ~. 2020.
\newblock Learning dynamic knowledge graphs to generalize on text-based games.
\newblock In \emph{NeurIPS 2020}.

\bibitem[{Ammanabrolu and Hausknecht(2020)}]{ammanabrolu2020graph}
Prithviraj Ammanabrolu and Matthew~J. Hausknecht. 2020.
\newblock \href {https://openreview.net/forum?id=B1x6w0EtwH} {Graph constrained
  reinforcement learning for natural language action spaces}.
\newblock In \emph{8th International Conference on Learning Representations,
  {ICLR} 2020, Addis Ababa, Ethiopia, April 26-30, 2020}. OpenReview.net.

\bibitem[{Ammanabrolu et~al.(2020)Ammanabrolu, Tien, Luo, and
  Riedl}]{ammanabrolu2020avoid}
Prithviraj Ammanabrolu, Ethan Tien, Zhaochen Luo, and Mark~O Riedl. 2020.
\newblock How to avoid being eaten by a grue: Exploration strategies for
  text-adventure agents.
\newblock \emph{arXiv preprint arXiv:2002.08795}.

\bibitem[{Cho et~al.(2014)Cho, van Merri{\"e}nboer, Bahdanau, and
  Bengio}]{cho-etal-2014-properties}
Kyunghyun Cho, Bart van Merri{\"e}nboer, Dzmitry Bahdanau, and Yoshua Bengio.
  2014.
\newblock \href {https://doi.org/10.3115/v1/W14-4012} {On the properties of
  neural machine translation: Encoder{--}decoder approaches}.
\newblock In \emph{Proceedings of {SSST}-8, Eighth Workshop on Syntax,
  Semantics and Structure in Statistical Translation}, pages 103--111, Doha,
  Qatar. Association for Computational Linguistics.

\bibitem[{Fulda et~al.(2017)Fulda, Ricks, Murdoch, and Wingate}]{fulda17}
Nancy Fulda, Daniel Ricks, Ben Murdoch, and David Wingate. 2017.
\newblock \href {https://doi.org/10.24963/ijcai.2017/144} {What can you do with
  a rock? affordance extraction via word embeddings}.
\newblock In \emph{Proceedings of the Twenty-Sixth International Joint
  Conference on Artificial Intelligence, {IJCAI} 2017, Melbourne, Australia,
  August 19-25, 2017}, pages 1039--1045. ijcai.org.

\bibitem[{Guo et~al.(2020)Guo, Yu, Gao, Gan, Campbell, and
  Chang}]{guo2020interactive}
Xiaoxiao Guo, Mo~Yu, Yupeng Gao, Chuang Gan, Murray Campbell, and Shiyu Chang.
  2020.
\newblock \href {https://doi.org/10.18653/v1/2020.emnlp-main.624} {Interactive
  fiction game playing as multi-paragraph reading comprehension with
  reinforcement learning}.
\newblock In \emph{Proceedings of the 2020 Conference on Empirical Methods in
  Natural Language Processing (EMNLP)}, pages 7755--7765, Online. Association
  for Computational Linguistics.

\bibitem[{Hausknecht et~al.(2019)Hausknecht, Loynd, Yang, Swaminathan, and
  Williams}]{hausknecht2019nail}
Matthew Hausknecht, Ricky Loynd, Greg Yang, Adith Swaminathan, and Jason~D
  Williams. 2019.
\newblock Nail: A general interactive fiction agent.
\newblock \emph{arXiv preprint arXiv:1902.04259}.

\bibitem[{Hausknecht et~al.(2020)Hausknecht, Ammanabrolu, C{\^{o}}t{\'{e}}, and
  Yuan}]{hausknecht19colossal}
Matthew~J. Hausknecht, Prithviraj Ammanabrolu, Marc{-}Alexandre
  C{\^{o}}t{\'{e}}, and Xingdi Yuan. 2020.
\newblock \href {https://aaai.org/ojs/index.php/AAAI/article/view/6297}
  {Interactive fiction games: {A} colossal adventure}.
\newblock In \emph{The Thirty-Fourth {AAAI} Conference on Artificial
  Intelligence, {AAAI} 2020, The Thirty-Second Innovative Applications of
  Artificial Intelligence Conference, {IAAI} 2020, The Tenth {AAAI} Symposium
  on Educational Advances in Artificial Intelligence, {EAAI} 2020, New York,
  NY, USA, February 7-12, 2020}, pages 7903--7910. {AAAI} Press.

\bibitem[{He et~al.(2016)He, Chen, He, Gao, Li, Deng, and
  Ostendorf}]{he2015deep}
Ji~He, Jianshu Chen, Xiaodong He, Jianfeng Gao, Lihong Li, Li~Deng, and Mari
  Ostendorf. 2016.
\newblock \href {https://doi.org/10.18653/v1/P16-1153} {Deep reinforcement
  learning with a natural language action space}.
\newblock In \emph{Proceedings of the 54th Annual Meeting of the Association
  for Computational Linguistics (Volume 1: Long Papers)}, pages 1621--1630,
  Berlin, Germany. Association for Computational Linguistics.

\bibitem[{Maaten and Hinton(2008)}]{maaten2008visualizing}
Laurens van~der Maaten and Geoffrey Hinton. 2008.
\newblock Visualizing data using t-sne.
\newblock \emph{Journal of machine learning research}, 9(Nov):2579--2605.

\bibitem[{Narasimhan et~al.(2015)Narasimhan, Kulkarni, and
  Barzilay}]{narasimhan15}
Karthik Narasimhan, Tejas Kulkarni, and Regina Barzilay. 2015.
\newblock \href {https://doi.org/10.18653/v1/D15-1001} {Language understanding
  for text-based games using deep reinforcement learning}.
\newblock In \emph{Proceedings of the 2015 Conference on Empirical Methods in
  Natural Language Processing}, pages 1--11, Lisbon, Portugal. Association for
  Computational Linguistics.

\bibitem[{Pathak et~al.(2017)Pathak, Agrawal, Efros, and
  Darrell}]{pathakICMl17curiosity}
Deepak Pathak, Pulkit Agrawal, Alexei~A. Efros, and Trevor Darrell. 2017.
\newblock \href {http://proceedings.mlr.press/v70/pathak17a.html}
  {Curiosity-driven exploration by self-supervised prediction}.
\newblock In \emph{Proceedings of the 34th International Conference on Machine
  Learning, {ICML} 2017, Sydney, NSW, Australia, 6-11 August 2017}, volume~70
  of \emph{Proceedings of Machine Learning Research}, pages 2778--2787. {PMLR}.

\bibitem[{Yao et~al.(2020)Yao, Rao, Hausknecht, and Narasimhan}]{yao2020calm}
Shunyu Yao, Rohan Rao, Matthew Hausknecht, and Karthik Narasimhan. 2020.
\newblock \href {https://doi.org/10.18653/v1/2020.emnlp-main.704} {Keep {CALM}
  and explore: Language models for action generation in text-based games}.
\newblock In \emph{Proceedings of the 2020 Conference on Empirical Methods in
  Natural Language Processing (EMNLP)}, pages 8736--8754, Online. Association
  for Computational Linguistics.

\bibitem[{Yu et~al.(2020)Yu, Guo, Feng, Zhu, Greenspan, and
  Campbell}]{yu2020deriving}
Mo~Yu, Xiaoxiao Guo, Yufei Feng, Xiaodan Zhu, Michael Greenspan, and Murray
  Campbell. 2020.
\newblock Deriving commonsense inference tasks from interactive fictions.
\newblock \emph{arXiv preprint arXiv:2010.09788}.

\end{thebibliography}
